\pdfoutput=1

\documentclass[11pt]{article}
\usepackage[table,xcdraw]{xcolor}
\usepackage{color}
\usepackage{acl}

\usepackage{times}
\usepackage{latexsym}
\usepackage{float}

\usepackage[T1]{fontenc}

\usepackage[utf8]{inputenc}

\usepackage{microtype}
\usepackage{graphicx}
\usepackage{booktabs}
\usepackage{multirow}

\title{Does Recommend-Revise Produce Reliable Annotations? \\ An Analysis on Missing Instances in DocRED}

\author{
    Quzhe Huang$^{1}$, 
    Shibo Hao$^{1}$, 
    Yuan Ye$^{1}$,
    Shengqi Zhu$^{2}$,
    Yansong Feng$^{1}$\thanks{\quad Corresponding author.}\and
    Dongyan Zhao$^{1}$ \\
    $^1$Wangxuan Institute of Computer Technology, Peking University, China \\ $^2$University of Washington \\
    {\tt \{huangquzhe,haoshibo,pkuyeyuan,fengyansong,zhaody\}} 
     {\tt @pku.edu.cn} \\
     {\tt sqzhu@uw.edu}
}

\begin{document}
\maketitle

\begin{abstract}
DocRED is a widely used 
dataset for document-level relation extraction. In the large-scale annotation, a \textit{recommend-revise} scheme is adopted to reduce the workload. Within this scheme, annotators are provided with candidate relation instances from distant supervision, and they then manually supplement and remove relational facts based on the recommendations. However, when comparing DocRED with a subset relabeled from scratch, we find that this scheme results in a considerable amount of false negative samples and an obvious bias towards popular entities and relations. Furthermore, we observe that the models trained on DocRED have low recall on our relabeled dataset and inherit the same bias in the training data.
Through the analysis of annotators' behaviors, we figure out the underlying reason for the problems above:
the scheme actually discourages annotators from supplementing adequate instances in the revision phase. We appeal to future research to take into consideration the issues with the recommend-revise scheme when designing new models and annotation schemes. The relabeled dataset is released at \url{https://github.com/AndrewZhe/Revisit-DocRED}, to serve as a more reliable test set of document RE models.

\end{abstract}

\section{Introduction}
Relation Extraction (RE) is an important task 
which aims to identify relationships held between entities in a given piece of text. While most previous methods focus on extracting relations from a single sentence \cite{lin2016neural, zhang-etal-2018-graph}, recent studies begin to explore RE at document level \cite{peng2017,gain, Latent,three,docunet}, which is more challenging as it often requires reasoning across multiple sentences.

The rapid development of document-level RE in the past two years has benefited from the proposal of DocRED \cite{DocRED}, the first large-scale and human-annotated dataset for this task. Noticeably, longer documents introduce an unprecedented difficulty in annotating the relation instances: as the total number of entities dramatically increases in accordance to text length, the expected number of \textit{entity pairs} surges \textit{quadratically}, intensively increasing the workload to check relationships between every pair. To address this problem, \citet{DocRED} applies a \textit{recommend-revise} process: in the \textit{recommendation} phase, a small set of candidate relation instances is generated through distant supervision; then, annotators are required to \textit{revise} the candidate set, removing the incorrect relation instances and supplementing the instances  not identified in the recommendation phase.

Shifting the construction process from scratch to an edit-based task, it seems that the recommend-revise scheme cuts down the effort of annotating by a large margin. However, whether the \textit{quality} of the annotation maintains a reliable standard in practice remains in doubt. To what extent can the accuracy of annotation be sacrificed due to the automated recommendation? And, how does the provided recommendation affect the behaviours of the annotators in the revision phase? Moreover, what are the real effects on the models trained on a dataset annotated with this scheme?

To answer these questions, we aim to provide a thorough comparison between careful annotations from scratch and the annotations under the recommend-revise scheme. We randomly select 96 documents from DocRED and ask two experts to relabel them from scratch independently. After annotating, the two experts come to a consensus of gold labels via discussion.
This revised dataset is publicly available at \url{https://github.com/AndrewZhe/Revisit-DocRED}, and we hope it can be used to evaluate the model's performance on real data distribution\footnote{While we cannot guarantee that the relabeled data
is totally error-free, we believe the quality is high enough to be approximated as a real distribution because each entity pair is examined by two annotators.}. With the help of these annotations, we discovered three sobering issues regarding the effects of the recommend-revise scheme:

\textbf{(1) A noticeable portion of relation instances is left out, and the distributional bias in the recommendation output is inherited, even after the revision process.} 
It is not surprising that recommendations alone fail to recognize all the relation instances, since RE models are far from perfect. Ideally, these unidentified instances should be added by human annotators during the revision phase. However, it turns out that $95.7\%$ of these missing instances are still left out even after revision. Furthermore, while the recommendations from distant supervision favor instances associated with popular entities and relations in the source Knowledge Base (Wikidata), this bias is still maintained and inherited even after human revision, leaving less popular relations and entities to be neglected. 

\textbf{(2) Worryingly, we find the models trained on DocRED have low recall on our relabeled dataset and they also inherit the same bias towards popular relations and entities.} We train recent models on DocRED and test them with the dataset relabeled by us. We notice that all models have much lower recalls on our dataset than previously reported on DocRED due to the numerous false negatives in training data, and those models are also biased to popular entities and relations. Further investigation reveals that the models' bias comes from the training set by comparing different strategies of negative sampling. 
Since one straightforward real-world application of relation extraction is to acquire novel knowledge from text, a RE model would be much less useful if it has a low recall, or perform poorly on less popular entities and relations.

\textbf{(3) The recommendations actually also impacts the behaviors of annotators, making them unlikely to supplement the instances left out.} 
This is the underlying reason for the two concerns above. We argue that the revision process fails to reach its goal, since it puts the annotators in a dilemma: while they are supposed to ``add” new instances left out by the recommendations, finding these missing instances may force the annotators to thoroughly check out the entities pair-by-pair, which is time-consuming and against the goal of this scheme. As a result, annotators can hardly make effective supplementation and would tend to perform the easier goal of \textit{validating existing relation instances}.

\section{Recommend-Revise Annotation Scheme}

The major challenge for annotating document-level RE datasets comes from the quadratic number of potential entity pairs with regard to the total number of entities in a document. As reported by \citet{DocRED}, a document in DocRED contains 19.5 entities on average, thus rendering 360 entity pairs with potential relationships. Therefore, for the 5,053 documents to be annotated, around 1,823,000 entity pairs are to be checked. Such workload will be around 14 times more than TACRED \cite{tacred}, the biggest human-labeled sentence-level RE dataset. Therefore, exhaustively labeling relations between each entity pair involves intensive workload and does not seem feasible for document-level RE datasets.

To alleviate the huge burden of manual labeling, \citet{DocRED} divides the annotation task into two steps: \textit{recommendation} and \textit{revision}. First, in the \textit{recommendation} phase, \citet{DocRED} takes advantage of Wikidata \cite{vrandevcic2014wikidata} and an off-the-shelf RE model to collect all the possible relations between any two entities in the same document. This process is automated and does not require human involvement. Then, during the \textit{revision} phase, the relations that exist in Wikidata or are inferred by the RE model for a specific entity pair will be shown to the annotators. Rather than annotating each entity pair from scratch, the annotators are required to review the recommendations, remove the incorrect triples and supplement the missing ones.

\section{Dataset}

\paragraph{DocRED}
The \textit{Document-Level Relation Extraction Dataset} (DocRED), introduced by \citet{DocRED}, is one of the largest and most widely used dataset for document-level relation extraction. 
DocRED consists of 5,053 English Wikipedia documents, each containing 19.5 entities on average. Every entity pair within a document may have one of the 96 types of relations or no relations, i.e., the additional \textit{no\_relation} label for negative instances.
In order to explore the supplementation in the revision phase and the influence of it on the released dataset, we acquire the original recommendations generated by distant supervision from the authors of DocRED. As we focus on the effect of missing instances, we do not consider the samples removed during the revision phase. 
The remaining annotations in the recommendations that are not removed later are denoted as $\mathbf{D_{Recommend}}$, and the annotations after human revision are denoted as $\mathbf{D_{Revise}}$.

\paragraph{DocRED from scratch}
To analyze the effect of the recommend-revise scheme, we re-annotate a subset of the documents used in DocRED from scratch and compare it with $\mathbf{D_{Recommend}}$ and $\mathbf{D_{Revise}}$. We randomly select 96 documents from the validation set of DocRED,  and each document is assigned to two experts to be annotated independently. They are explicitly required to check every entity pair in the documents and decide the relationships entirely based on the original text with no recommendation. This turns out to be an extraordinarily difficult task where each document takes up half an hour for annotation on average.
The inter-annotator Cohen's Kappa is 0.68 between our two experts, indicating a high annotation quality. 
After that, the two experts discuss the inconsistent instances together and reach an agreement on the final labels. 
As this paper focuses on the bias caused by \textit{false negatives} in the recommend-revise scheme, we assume the labeled instances in DocRED are all correct.

For the instances labeled in DocRED but not by our experts, we add them to our annotation. We denote this new annotation set as {$\mathbf{D_{Scratch}}$}.

\section{Dataset Comparison}

\begin{table*}[]
\small
    \centering
\begin{tabular}{lrllcc}
\toprule
{} &  \# Instance &     \# Pop Rel &   \# Unpop Rel & $popularity_{max}$ & $popularity_{min}$ \\
\midrule
$\mathbf{D_{Recommend}}$               &        1167 &   659 (56.5\%) &   508 (43.5\%) &              294.4 &               85.2 \\
$\mathbf{D_{Revise}}$                  &        1214 &   676 (55.7\%) &   538 (44.3\%) &              291.5 &               84.4 \\
$\mathbf{D_{Scratch}}$                 &        3308 &  1615 (48.8\%) &  1693 (51.2\%) &              266.3 &               67.4 \\
$\mathbf{D_{Revise} - D_{Recommend}}$  &          47 &    17 (36.2\%) &    30 (63.8\%) &              221.3 &               66.0 \\
$\mathbf{D_{Scratch} - D_{Recommend}}$ &        2141 &   956 (44.7\%) &  1185 (55.3\%) &              251.0 &               57.7 \\
$\mathbf{D_{Scratch} - D_{Revise}}$    &        2094 &   939 (44.8\%) &  1155 (55.2\%) &              251.7 &               57.5 \\
\bottomrule
\end{tabular}
    \caption{Statistics of datasets. \#Instance means the total relation instance in the dataset. \# Pop Rel and \# Unpop Rel shows the number of instances associated with popular relations and unpopular relations respectively. The last two columns represent the average entity popularity across all the relation instances, with the \textit{popularity}$_{max}$ indicating the higher popularity of head and tail entities in an instance, and \textit{popularity}$_{min}$ indicating the lower one.}
    \label{tab:dataset_statistics}
\end{table*}

Table~\ref{tab:dataset_statistics} shows the statistics and comparison of $\mathbf{D_{Scratch}}$, $\mathbf{D_{Recommend}}$ and $\mathbf{D_{Revise}}$ on the 96 randomly-selected documents in DocRED.

\subsection{False Negatives in Recommendation}
\label{subsec:false_neg}
Comparing $\mathbf{D_{Recommend}}$ with $\mathbf{D_{Scratch}}$, it is noticeable that huge amounts of ground-truth annotation labels are left out. While $\mathbf{D_{Recommend}}$ captures 1167 relation instances in the documents, a more careful, entity-by-entity examination as did in $\mathbf{D_{Scratch}}$ would reveal that there are as much as 3308 relation instances within the same documents. This shocking fact reveals that almost two-thirds of the relation instances are missing and wrongly labeled as negative.

Another unexpected fact is that annotators hardly added anything during the revision phase. The final version reports 1214 relation instances, with a mere increase of 47 (1.4\%) cases in total, or 0.49 instances on average for each document. This suggests that while we had great hopes of our revision process to make things right, it is not working to a sensible extent: the majority of the unlabeled instances, which take up nearly two-thirds of the instances, simply remain out there as they were.

\subsection{Dataset Bias}
Given the analysis above, another even more serious issue arises: since the changes introduced by the revision are so limited, the output after revision may still contain the same bias as in the recommendation. That is, if the recommendations contain a systematic flaw, the new dataset will probably keep on inheriting it. In this section, we verify that such biases largely exist in the recommendation phase and are thus inherited to the DocRED dataset.

The recommendations of DocRED are collected from two sources: Wikidata and a relation extraction model. However, if we consider the facts reserved after revision by annotators, where wrongly labeled ones get removed, the majority of them are taken directly from Wikidata\footnote{See Appendix \ref{sec:appendix_source} for details.}.

We suggest that as a collaborative knowledge base, the relation instances related to common entities and properties are more likely to be collected and added to Wikidata. In such cases, the recommendation from Wikidata will naturally favor popular entities and relations, while the less common ones would be left out. We validate this hypothesis in the following sections, where we investigate the bias of DocRED from the perspective of both relations and entities.

\subsubsection{Bias of Relations}

To determine whether the data set has a preference for popular relationships, we divide the 96 relationships in DocRED into two categories using Wikidata statistics and then compute their distribution. Specifically, we acquire the \textit{List of top 100 properties by quantity of item pages that link to them} from Wikidata's official website\footnote{\url{https://www.wikidata.org/wiki/wikidata:Database_reports/
List_of_properties/Top100}} and consider a relation as popular if it appears on this list. Among the 96 relationships in DocRED, 25 are in top 100, including \textit{country}, \textit{publication date}, and so on.

The center two columns of Table~\ref{tab:dataset_statistics} illustrate the distribution of these two categories of relationships across multiple datasets. First, we can see that in the real distribution, i.e., in $\mathbf{D_{Scratch}}$, the percentages of these two types of relations are 48.8\% and 51.2\%, respectively, which is close to 1:1 with slightly \textit{fewer} popular relations. However, the proportion of all instances belonging to the popular relationship reached 56.5\% in recommendations, $\mathbf{D_{Recommend}}$, which is significantly higher than the 43.5\% for unpopular ones. Further study of those instances that were mistakenly excluded during the recommendation phase, $\mathbf{D_{Scratch} - D_{Recommend}}$, reveals that cases involving unpopular relationships are more likely to be missing. This demonstrates that the recommendation phase in DocRED does have a systematic bias related to the popularity of relations.
The instances supplemented during the revision phase, $\mathbf{D_{Revise} - D_{Recommend}}$, help to mitigate this bias marginally, whereas annotators label more instances belonging to unpopular relations. However, in comparison to $\mathbf{D_{Scratch}}$, which represents the real relation distribution, $\mathbf{D_{Revise}}$ still prefers popular relations. This is because the annotators place an excessive amount of trust in the recommendations and do not add sufficient missing instances during the revision phase. According to the statistics in the Table~\ref{tab:dataset_statistics}, the recommendation's bias toward the relation is ultimately inherited by the dataset that passed manual inspection.

\subsubsection{Bias of Entities}
We hypothesize that the instances involving very popular entities are more likely to appear in Wikidata recommendations, whereas instances related to extremely rare entities are more likely to be disregarded.
To determine whether such bias exists, we analyze the popularity of entities engaged in relation instances across multiple data sets.
Each named entity in DocRED is linked with a Wikidata item based on the literal matching of names or aliases\footnote{Items in Wikidata may have multiple \textit{aliases}; we say an item is matched with an entity if any one of the aliases is the same with the entity name.}. The popularity of an entity is represented by how many times the matched item appears in a relation instance in Wikidata (either as head or tail); if an entity matches more than one Wikidata items, the highest count among the matched items is taken as its popularity. For those entities that cannot be linked to Wikidata, we assign a popularity of -1.

For each relation instance, we compute two types of popularities. Since an instance contains a pair of entities (head and tail) usually with different popularities, we define \textit{popularity}$_{max}$ to be the higher popularity of the pair of entities, and \textit{popularity}$_{min}$ to be the lower one. We report the average popularity of relation instances in each dataset in Table~\ref{tab:dataset_statistics}.

Comparing $\mathbf{D_{Recommend}}$ and $\mathbf{D_{Scratch}}$, we find that the former's \textit{popularity}$_{max}$ is 294.4, far more than the latter's 266.3. This means that instances containing popular entities will be more likely to be retained during the recommendation phase. Regarding those instances that were incorrectly excluded during the recommendation phase, $\mathbf{D_{Scratch} - D_{Recommend}}$, their \textit{popularity}$_{min}$ is 57.7, which is less than the 67.4 in $\mathbf{D_{Scratch}}$. This demonstrates that instances involving uncommon entities are more likely to be ignored during the recommendation phase.

This entity-related bias is apparent in the revised data set as well. The \textit{popularity}$_{max}$ kept by $\mathbf{D_{Revise}}$ remains larger than that of $\mathbf{D_{Scratch}}$, while the \textit{popularity}$_{min}$ of $\mathbf{D_{Scratch} - D_{Revise}}$ is also lower than that of $\mathbf{D_{Scratch}}$. This is mostly because the facts supplemented at the revision phase is too few to eliminate such bias.

\section{Model Bias}

To investigate if RE models trained on such data will likewise learn the same bias, we train and select RE models on the recommend-scheme-labeled dataset, $\mathbf{D_{Revise}^{Train}}$ and $\mathbf{D_{Revise}^{Valid}}$ and then assess the models' performance on the real data distribution, $\mathbf{D_{Scratch}}$. The construction process of $\mathbf{D_{Revise}^{Train}}$ and $\mathbf{D_{Revise}^{Valid}}$ is the same as $\mathbf{D_{Revise}}$, while the former is actually the original train set and the latter is the validation set in DocRED excluding the 96 documents in $\mathbf{D_{Revise}}$.
In those settings, we examine the performance of recent models: BiLSTM \cite{DocRED}, GAIN-BERT$_{base}$ \cite{gain}, SSAN-Roberta$_{large}$ \cite{ssan}, ATLOP-Roberta$_{large}$ \cite{atlop} and DocuNet-Roberta$_{large}$ \cite{docunet}. The last three models are the most competitive ones for DocRED currently, while the others are shown to make sure that our analysis can generalize to models of smaller sizes.

\subsection{Overall Performance}
\begin{table}[]
\small
\begin{tabular}{lccc|ccc}

\toprule
{} & \multicolumn{3}{c|}{Revise} & \multicolumn{3}{c}{Scratch} \\
{} &   P &   R &  F1 &   P &   R &  F1 \\
\midrule
BiLSTM  &  50.2 &  46.7 &  48.4 &  66.6 &  22.8 &  33.9 \\
GAIN    &  60.0 &  56.8 &  58.3 &  81.1 &  28.1 &  41.8 \\
ATLOP   &  66.3 &  59.1 &  62.5 &  90.3 &  29.5 &  44.5 \\
SSAN    &  63.1 &  61.3 &  62.2 &  84.5 &  30.1 &  44.4 \\
DocuNet &  66.9 &  59.9 &  63.2 &  89.1 &  29.3 &  44.1 \\
\bottomrule
\end{tabular}
\caption{Results for different RE models evaluated on $\mathbf{D_{Revise}}$ and $\mathbf{D_{Scratch}}$.}
\label{tab:diff}
\end{table}

Table~\ref{tab:diff} summarizes the evaluation results of five models on $\mathbf{D_{Revise}}$ and $\mathbf{D_{Scratch}}$. All results were reported using micro-average F1-scores as in prior literature \cite{gain,atlop}. 

Notably, we observe a significant decline in F1 for all the 5 models on $\mathbf{D_{Scratch}}$ which is mainly due to the dramatic drop in the recall.
The drop is the result of the bias in training data, i.e., the model trained on biased data lacks the generalization ability to extract relation instances that are systematically missed in the dataset. We will validate this point in the following section.

\subsection{Bias from Data to Model}
To better understand the different performances on the two datasets, we analyze the model capability
over different relations and entities. Not surprisingly, we find that models trained on $\mathbf{D_{Revise}^{Train}}$ prefer popular entities and relations as well. Additional experiments suggest that this may be because missing instances are considered as negative samples during training. Given that a substantial proportion of unlabelled instances are associated with unpopular entities and relations, the model will be forced to disregard those unpopular ones under the incorrect penalty for the missing instances.
\begin{figure}[pt]
    \centering
    \includegraphics[width=0.45\textwidth]{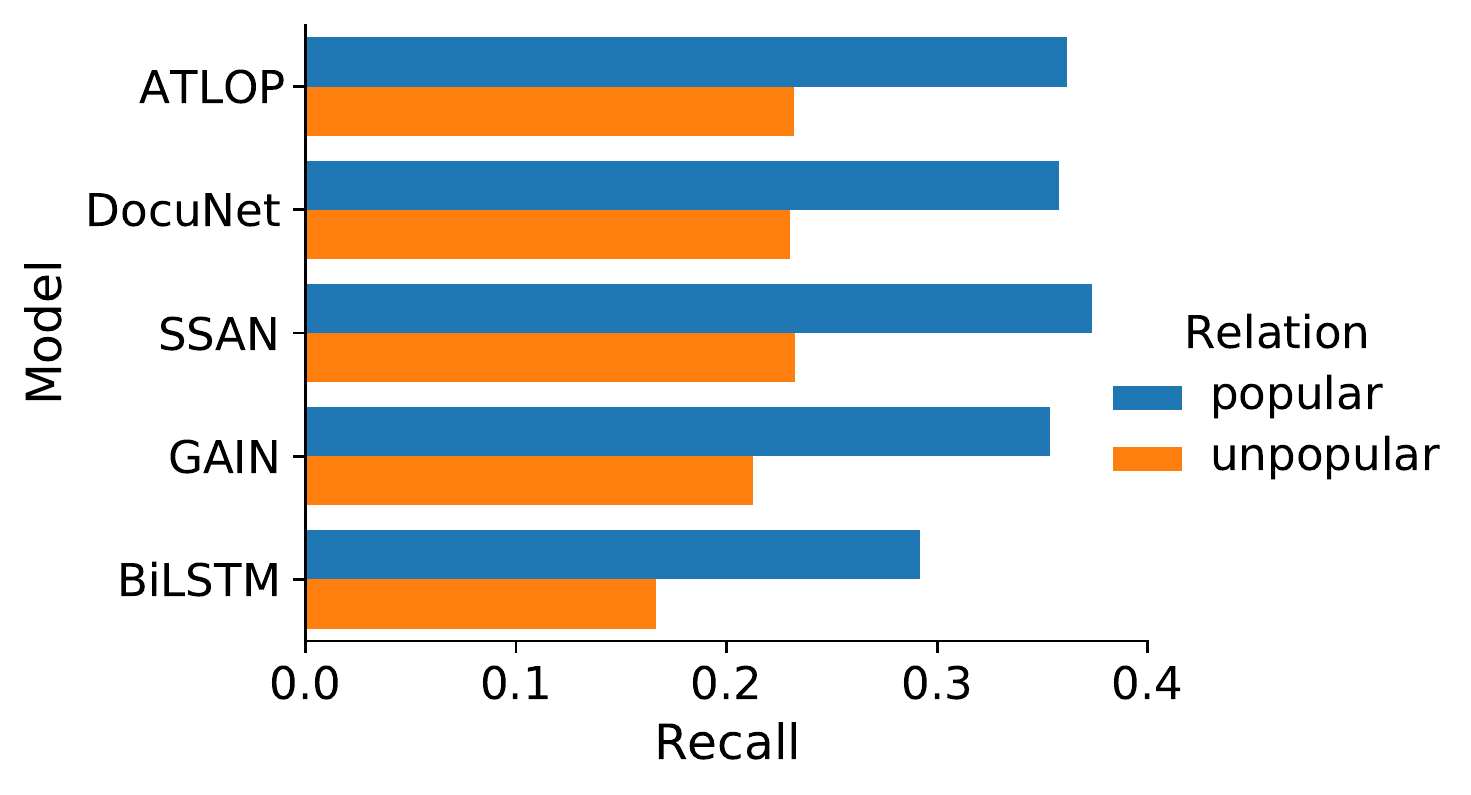}

    \caption{The recall of models on instances associated with popular and unpopular relations.
    }
    \label{fig:relation_bias}
\end{figure}
\paragraph{Relation Bias} Figure~\ref{fig:relation_bias} shows the recall of the models on the instances associated with popular and unpopular relations respectively. As is depicted, if an instance's relation is popular, it is almost twice as likely to be successfully extracted compared with an instance whose relation is not popular. This gap does not narrow with the improvement of the model's overall performance. The difference between the probability of successfully extracting popular and unpopular relations is 0.129 for the best model ATLOP, which is even greater than the 0.125 for BiLSTM. This indicates that all models trained on the original DocRED favor popular relations and ignore the unpopular ones.

\begin{figure}[pt]
    \centering
    \includegraphics[width=0.40\textwidth]{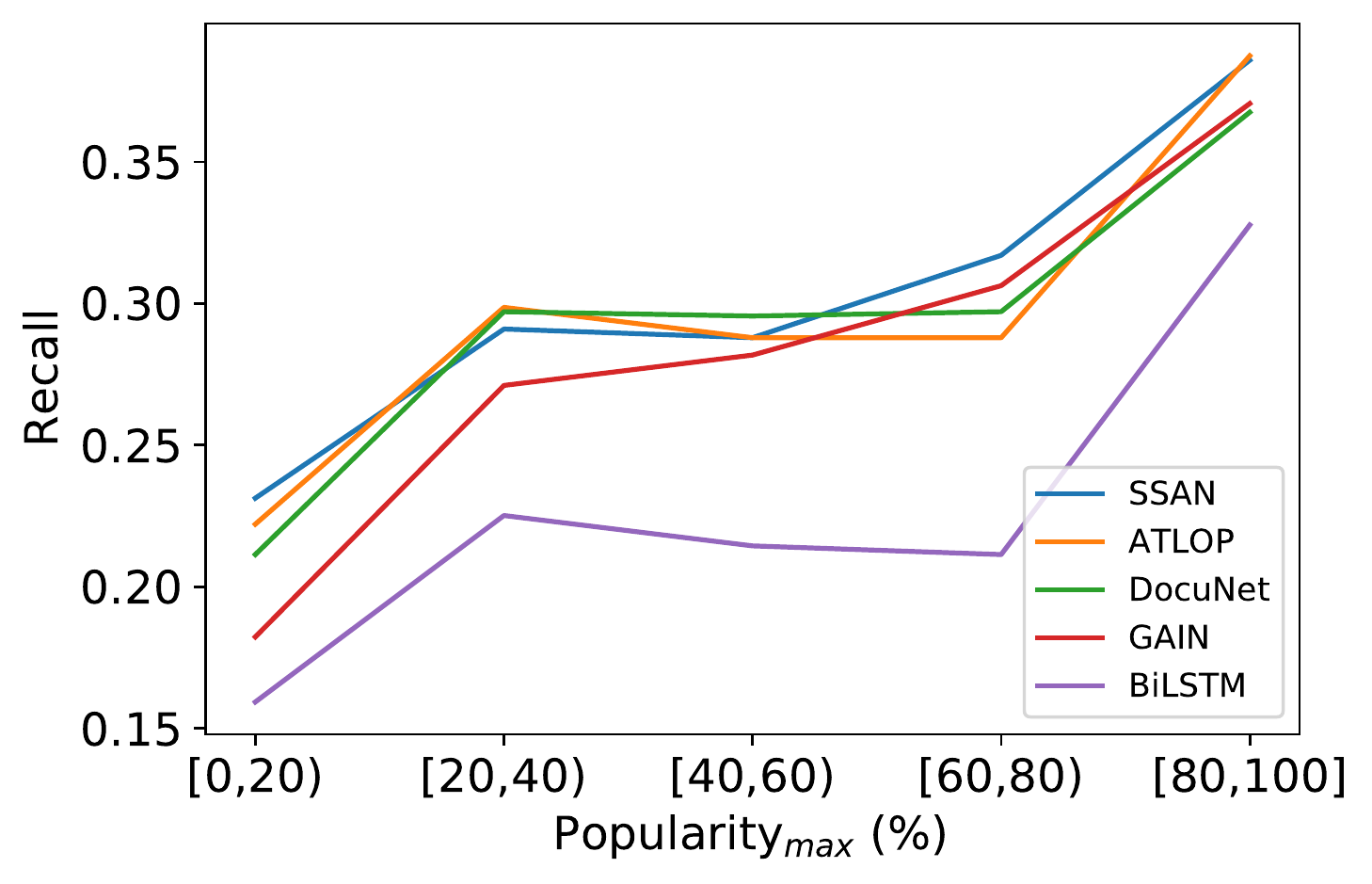}

    \caption{Model's recall in instances with entities of different popularity. We divide all instances in $\mathbf{D_{Scratch}}$ into 5 groups based on the \textit{popularity}$_{max}$ in each instance and we can see that the probability of discovering an instance is positively correlated with its \textit{popularity}$_{max}$.
    }
    \label{fig:entity_bias}
\end{figure}

\paragraph{Entity Bias} Figure~\ref{fig:entity_bias} shows the model's recall curve as the \textit{popularity}$_{max}$ of instances in $\mathbf{D_{Scratch}}$ increases\footnote{The curve with \textit{popularity}$_{min}$ is shown in Appendix \ref{sec:appendix_entity_bias}.}. We divide all instances in $\mathbf{D_{Scratch}}$ into 5 groups based on the $popularity_{max}$ in each instance, and we calculate the recall for each group independently. As seen in Figure~{\ref{fig:entity_bias}}, all the curves exhibit a clear rising trend, indicating that the probability of discovering an instance is positively correlated with its $popularity_{max}$. Additionally, we can see that the middle of the ATLOP's and DocuNET's curves is nearly horizontal, which means that they are more sensitive to extremely popular or particularly rare entities.

\subsection{Missing Instances as Negative Samples}
Previous works \cite{gain,atlop,docunet} regard any instances that are not annotated with any relations as label \textit{no\_relation}, which means the missing instances are treated as negative samples during training and a model will be punished for predicting them as positive. We thus hypothesize the model's bias originates from the incorrect penalty for missing instances in the training process. To demonstrate this, we generate the negative samples in a different approach, using the instances manually eliminated during the revision step only. We denote such construction of negative samples as
$\mathbf{N_{Hum}}$, and the method that treats all samples other than the positive instances as negative is called $\mathbf{N_{All}}$.
Due to the fact that the sample generated by $\mathbf{N_{Hum}}$ has been manually verified, there is no issue with false no\_relation instances. We train the same models using $\mathbf{D_{Revise}^{Train}}$ with negative samples constructed by $\mathbf{N_{Hum}}$ and $\mathbf{N_{All}}$ and compare models' preference for popular entities and relations.

\begin{figure}[pt]
    \centering
    \includegraphics[width=0.40\textwidth]{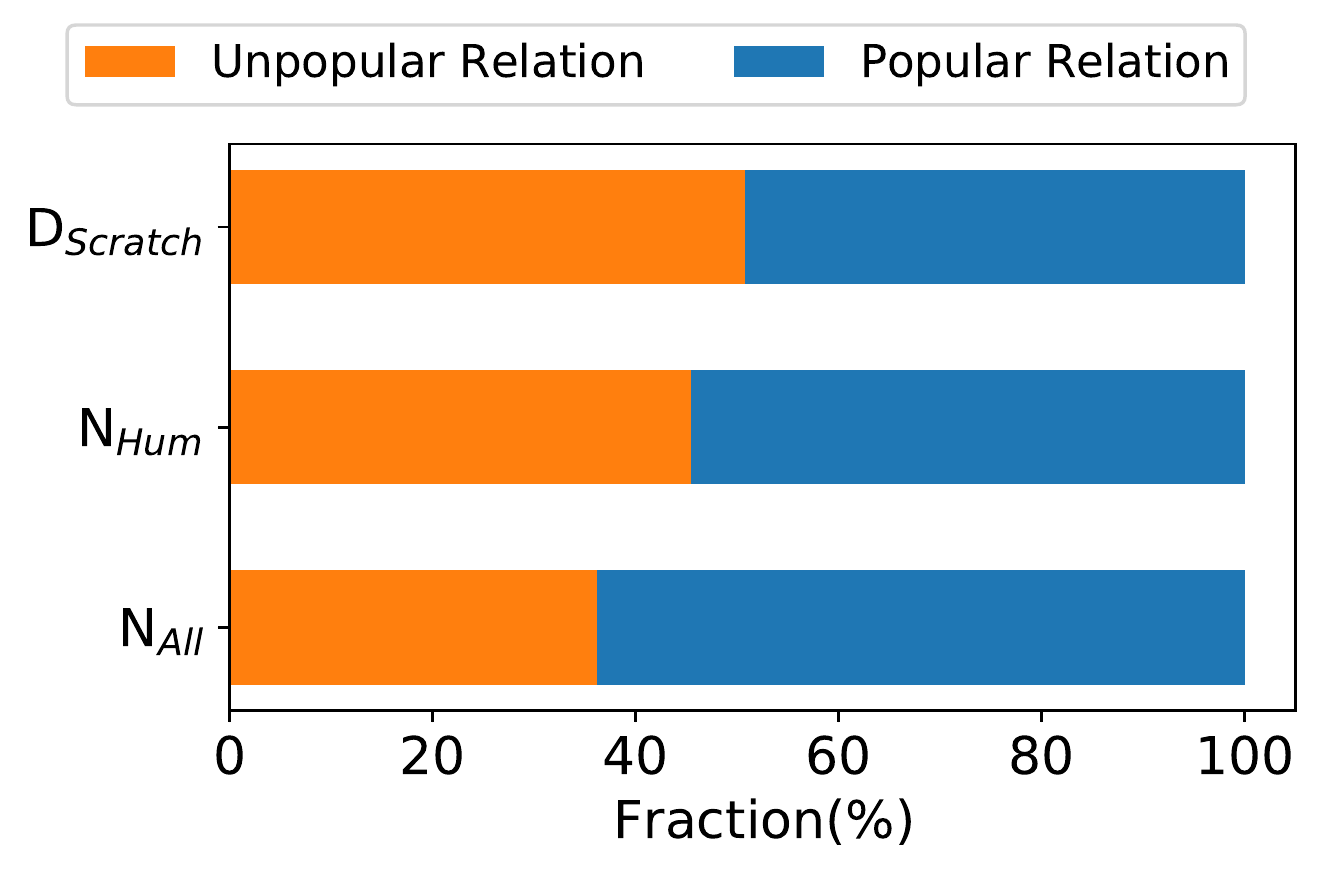}

    \caption{ The proportion of instances associated with popular and unpopular relationships in the correct prediction of GAIN. In comparison to N$_{All}$, the model trained with N$_{Hum}$ predicts more unpopular relation associated instances, and the ratio between popular and unpopular relationships is closer to the distribution in $\mathbf{D_{Scratch}}$.  
    }
    \label{fig:relation_bias_cause}
\end{figure}

Figure~\ref{fig:relation_bias_cause} depicts the fraction of instances that correspond to the popular relationship among the instances accurately predicted by GAIN trained with $\mathbf{D_{Revise}^{Train} + N_{Hum}}$ and $\mathbf{D_{Revise}^{Train} + N_{All}}$. Additionally, we mark  the true distribution of the data in $\mathbf{D_{Scratch}}$. As can be seen, when trained with $\mathbf{D_{Revise}^{Train} + N_{Hum}}$, GAIN can find more unpopular relation associated instances and the gap between the proportion of unpopular relation associated in model's prediction and $\mathbf{D_{Scratch}}$ is narrowed down.

\begin{figure}[pt]
    \centering
    \includegraphics[width=0.5\textwidth]{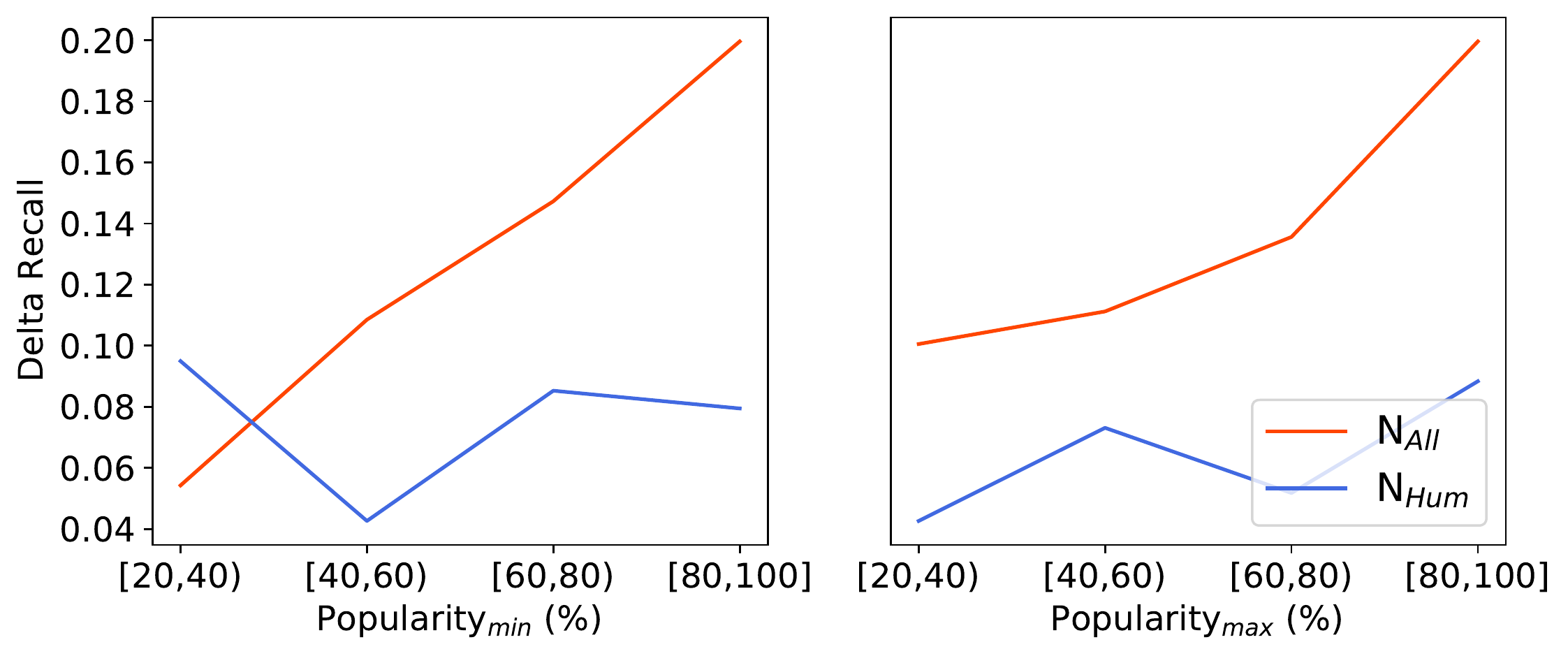}

    \caption{The gain of recall in instances with different \textit{popularity}$_{min}$ and  \textit{popularity}$_{max}$, compared with the [0-20) group.
    }
    \label{fig:entity_bias_cause}
\end{figure}

Based on the entity popularity in each instance, we partition all instances in $\mathbf{D_{Scratch}}$ into five categories and calculate the recall for each group independently. Figure~\ref{fig:entity_bias_cause} shows the improvement of GAIN's recall compared with the group which includes the instances with the most unpopular entities (0-20\%). In comparison to $\mathbf{N_{All}}$, using $\mathbf{N_{Hum}}$ to construct negative samples to train a model will dramatically lessen the rising trend of the model's recall as the entity's popularity grows. 

\section{Annotators’ Dilemma}

Finally, we move on to discuss another more implicit influence of the recommend-revise scheme on the \textit{annotators’} aspect. As discussed in Section \ref{subsec:false_neg}, while we expected the revision process to help supplement the instances left out, it turns out that an incredibly low number is added indeed.
Given that the annotators are trained to accomplish the revision task, we wonder why they still fail in such a uniform manner. 
We would like to argue that it is the nature of the revision process that puts the annotators in a dilemma, where they have to choose between a huge effort and insufficiency of supplementation. 

Recall that there is a distinct difference in the settings of \textit{examining} a labeled relationship and \textit{supplementing} an unidentified relationship. 
For the former, annotators are required to find evidence for a recommended relation instance and remove it if there is conflicting or no evidence. This process only requires checking a single entity pair and collecting the information related to the two specific entities.
However, this is not the case for \textit{supplementing} a possible, unidentified relation instance, which can exist between any entity pair. There is no clear range of searching or indicating information; all they can do is to check pair-by-pair, just like what they do from scratch. This puts annotators in an awkward dilemma, especially when they understand the motivation of this scheme: if they are to be fully responsible for the missing instances at large, they will always have to complete the thorough pairwise checking one by one; however, this would make the whole process of the recommend-revise scheme meaningless in return, as it's just like a practice from scratch.

The harsh requirements of supplementing push annotators to overly rely on the recommendation results and simply examine them. This is especially worth worrying about in real practice, where annotators are recruited to complete a certain number of annotations, and typically paid according to the estimated number of hours or the total number of instances they devote to the annotation\cite{draws2021checklist}.
Under this dilemma, it is a natural result that they are especially unmotivated to carry out the exhaustive checking for supplementation in order to get a reasonable pay in the given time.

\begin{figure}[t]
    \centering
    \includegraphics[width=0.48\textwidth]{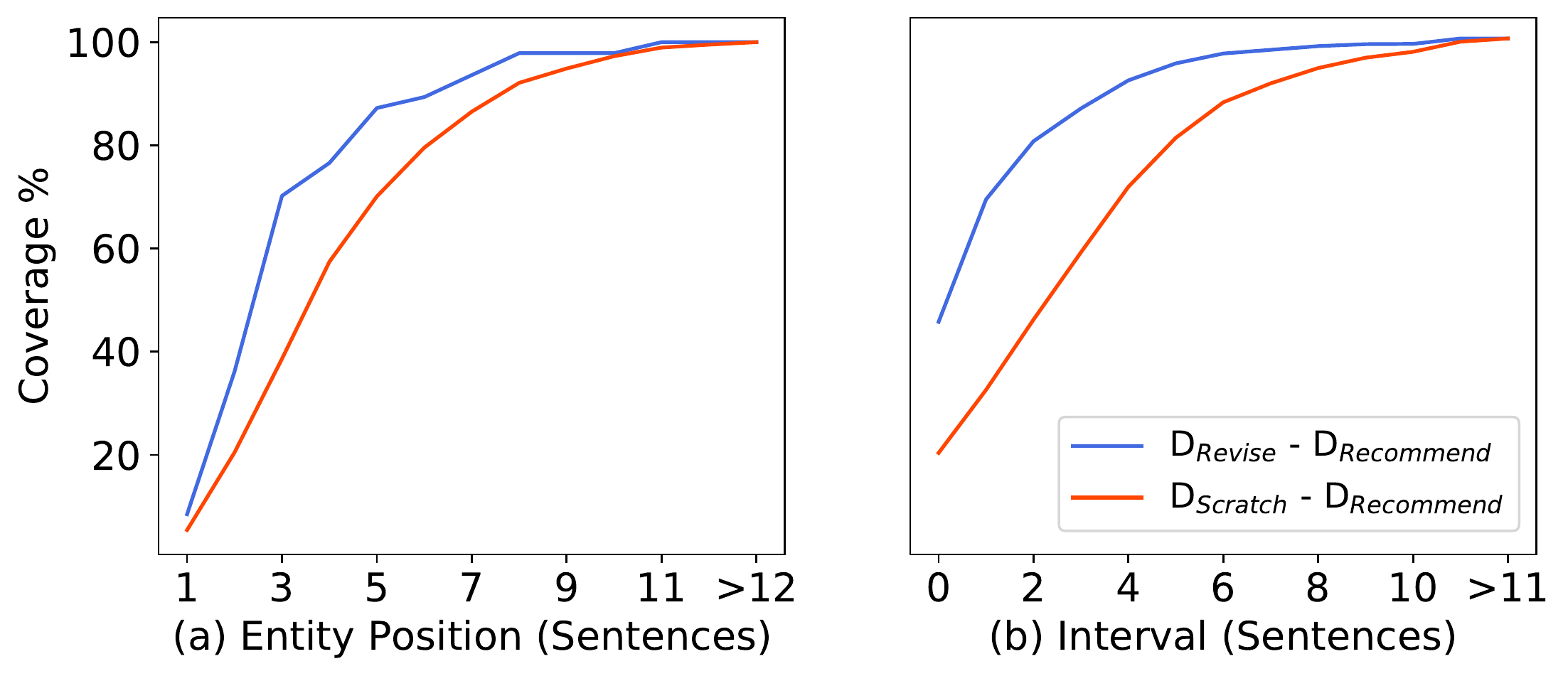}

    \caption{(a) illustrates the relations coverage measuring the position of head entity's first appearance.  (b) shows the coverage measuring the minimum distance in sentences between the mentions of the head and tail entities of a relation.
    }
    \label{fig:entity_analysis}
\end{figure}

In fact, we observe an interesting phenomenon that annotators largely tend to just pick some most obvious missing instances, convince themselves that they have accomplished the supplementation, and simply move on to the next document. This can be seen in Figure~\ref{fig:entity_analysis}, where we compare the distributional characteristics of the successfully supplemented instances ($\mathbf{D_{Revise}}$ - $\mathbf{D_{Recommend}}$) and all the missing instances in general ($\mathbf{D_{Scratch}}$ - $\mathbf{D_{Revise}}$). 
Sub-figure (a) shows the accumulative statistics of the position of the head entity's first appearance in the document. 
We can see that the instances added by annotators in DocRED exhibit an extremely obvious tendency to occur earlier in the text, where more than 70\% added instances are in the first 3 sentences. In contrast, all missing relation instances as a whole are almost distributed in every part of the document uniformly. This reveals the interesting fact that humans typically tend to pick up the relations where the entities in it are mentioned earlier in the document. Sub-figure (b) further compares the minimum distance between the mentions of the head and tail entities of one relation instance. We once again see the interesting fact that annotators have a strong tendency to add the ``most easily identifiable’’ instances where the head and tail entities are quite close. Specifically, the proportion of entity pairs mentioned in just one single sentence (Interval=0) is around 20\% for all missing facts, but is as high as 45\% for the ones chosen by annotators to be supplemented.
This tells us how annotators naturally avoid burdens of reading brought by longer intervals, which possibly indicates more complicated inference with multiple sentences.

\begin{figure*}[t]
    \centering
    \includegraphics[width=0.93\textwidth]{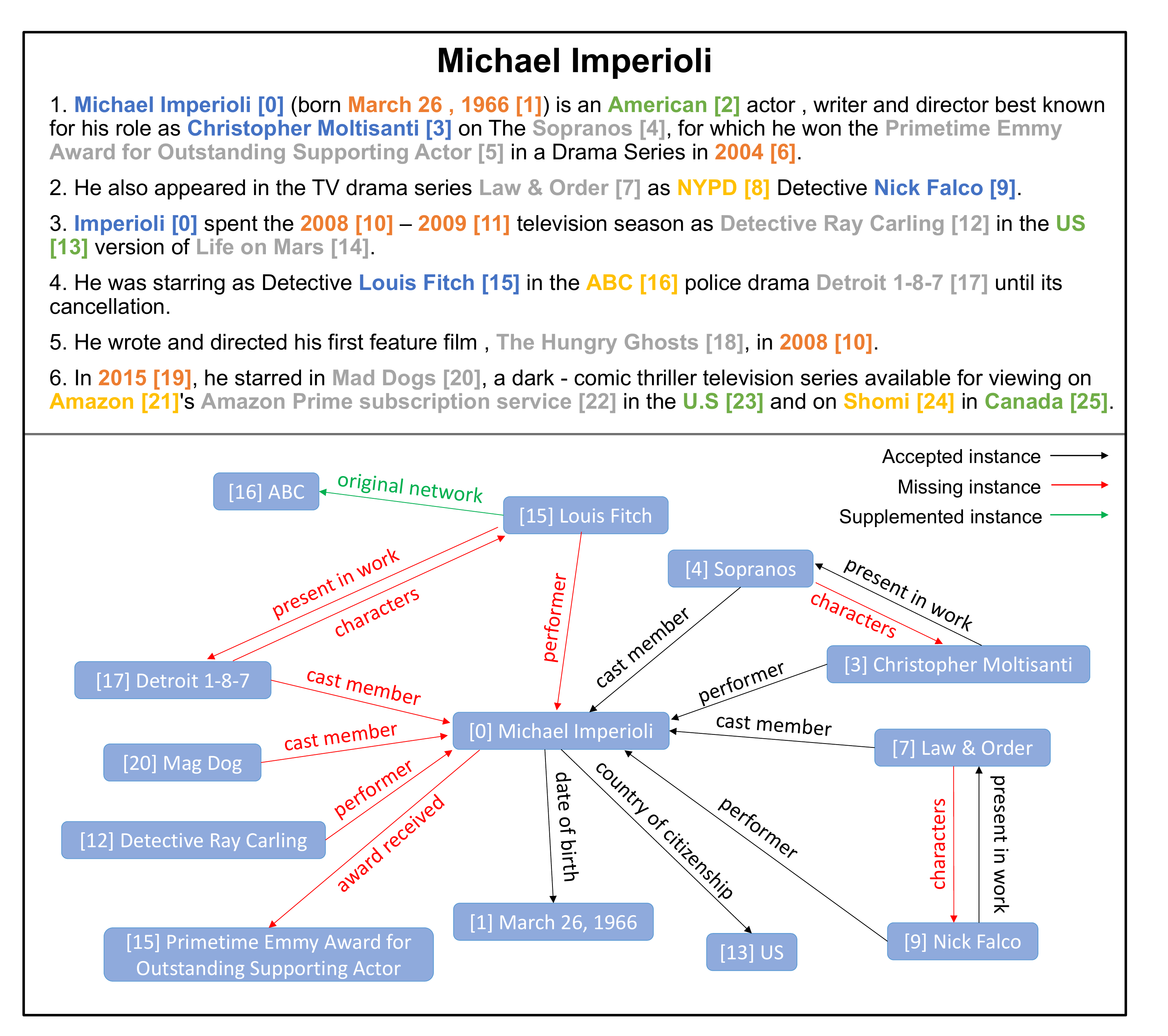}

    \caption{A case from DocRED. The upper part shows the original documents with entities highlighted according to their types (\textcolor[RGB]{68,114,196}{PER}, \textcolor[RGB]{237, 125, 49}{TIME}, \textcolor[RGB]{255, 192, 0}{ORG}, \textcolor[RGB]{112, 173, 71}{LOC}, \textcolor[RGB]{165, 165, 165}{MISC}). The lower part is an illustration of DocRED's annotation related to entity \textit{Michael Imperioli}, where the black arrows indicate the instances that are recommended and accepted by annotators, the red ones indicate those not recommended and eventually missed in the dataset, and the green ones indicate those not recommended but correctly supplemented by annotators. The instances that are recommended but rejected by annotators are not shown in the figure. 
    }
    \label{fig:case}
\end{figure*}

From these observations, we see that there exist clear patterns among the very few instances added by human annotators. This reveals a serious fact that annotators are intentional in ``pretending’’ to be supplementing with the least possible effort. Given the consensus behavior of annotators and the very limited number of additional, it is most likely that the nature of the annotation task pushes the annotators to this embarrassing dilemma of adding and abandoning. Thus, we propose a call to the NLP community that researchers should always be aware that annotation schemes, like the recommend-revise scheme, can have a direct impact on the annotation workers, affecting their willingness and behaviors, and thus have a deeper influence on the collected data.

\section{Case Study}

We can summarize all these problems mentioned above in the annotation with a concrete case in DocRED shown in Figure~\ref{fig:case}. 
The figure depicts the annotations associated with the entity \textit{Michael Imperi}, as well as the relation that is added in revision.
Let's first focus on the red edges, which indicate the relation triples that are neither recommended nor supplemented by human. Regrettably, half of the total 18 relation triples remain missing, and just one triple is added during revision (the green edge). Compared with black edges, which indicated correctly annotated instances, the red edges are more likely to be associated with less popular entities. For example, "Sopranos" [4] and "Law \& Order" [7], two popular series with at least 100K+ comments on IMDB, and about 200 edges in Wikidata, are connected with "Michael Imperioli" [0] with the relation "cast member" in the annotation, but "Detroit 1-8-7" [17] and "Mad Dogs" [20], supposed to hold the same relation to "Michael Imperioli" [0], are missed. In the text, all these series appear in similar circumstances, and the only difference is the latter ones are not recommended to the annotators, essentially because of their less popularity (less than 10K comments on IMDB, and less than 50 edges in Wikidata).
We can also see the effect on the popularity of relations in the connection between [7] and [9]. "present in work" and "characters" should occur symmetrically according to the definitions, but the latter one is missed in the recommendation. Correspondingly, in Wikidata, the latter relation has 19057 links, which is less than the former's 82250 links.
The last point to notice is the only green edge between Louis Fitch [15] and ABC [16], which is not recommended, but supplemented by annotators. Among all the missed instances in the recommendation, the annotators only supplement this one, which is easy to identify in the text due to both the head and tail entities being mentioned in the same sentence. This is consistent with our analysis of annotators' behavior above.

\section{Related Works}

With the advance of deep learning models, the annotation sometimes becomes the bottleneck for a machine learning system. Recently, analyzing the annotation quality 
has received increasing attention. \citet{pervasive} collects and analyzes the label errors in the test sets of several popular benchmarks, showing label errors are ubiquitous and destabilize machine learning benchmarks. More specific to RE task, \citet{Revisited} addresses the annotation problems in TACRED \cite{tacred}, a popular sentence-level RE dataset. They find label errors account for 8\% absolute F1 test error and more than 50\% of the examples need to be relabeled. \citet{Re-TACRED} expands this research to the whole dataset, resulting in a complete re-annotated version, Re-TACRED, and conducts thorough analysis on the models' performance. Our work differs from them in that we delve into the nature of document-level RE task, and especially explore how the error is systematically introduced into the dataset through \textit{recommend-revise} scheme.

Methodologies to solve incomplete annotations for information extraction tasks have been widely discussed in previous works. Different from classification tasks, information extraction requires annotators to actively retrieve positive samples from texts, instead of just assigning a label for a given text. The problem is also attributed to the use of distant supervision \cite{Mintz} where the linked KG is not perfect. Some works apply general approaches like positive unlabeled learning \cite{JRE_FN, NER_positive} or inference learning \cite{Inference}. Task-specific models are also designed, like Partial CRF \cite{PartialCRF} for NER \cite{Yang2018}, and novel paradigm for joint RE \cite{JRE_FN}. However, none of them examine the distribution bias in the training data, and those methods are not validated in the context of the document-level RE task.

Prevalent effective methods on document-level RE include graph-based models and transformer-based models. Graph-based models like \citet{gain} and \citet{docunet} are designed to conduct relational reasoning over the document, and transformer-based models \cite{atlop,ssan} are good at recognizing long-distance dependencies. However, all previous models treat unlabeled samples in the dataset as negative samples, and do not concern the problems in annotations. We believe our analysis and re-annotated dataset will help future work focus more on the discrepancy between the annotation and real-world distribution, instead of just overfitting the dataset.

\section{Conclusion}
In this paper, we show how the recommend-revise scheme for DocRED can cause bias and false negative issues in the annotated data. The flaws of dataset affect the model's recall on real data and also teach the model the same bias in training data. As this scheme cannot reduce the human labor essentially without the loss of annotation quality, more efficient strategies for annotation are to be explored. On the other hand, considering that building a reliable training set for document RE is extremely expensive, it is also a meaningful topic that how to alleviate the dataset shift problem \cite{shift} by injecting appropriate inductive bias into the model's structure, instead of inheriting the bias in the training data. We believe the in-depth analysis provided in this paper can benefit future designs of document-level RE models, and our \textit{Scratch} dataset can serve as a fairer test set.

\section*{Acknowledgements}
This work is supported in part by National Key R\&D Program of China (No. 2020AAA0106600) and NSFC (62161160339). We would like to thank the anonymous reviewers and action editors for their helpful comments and suggestions; thank Weiye Chen for providing feedback on an early draft. For any correspondence, please contact Yansong Feng.

\section*{Ethical Considerations}
This work focuses on quality checking and re-annotations of DocRED, a publicly available dataset constructed from Wikipedia pages. All source documents and types of relationships are provided and utilized in the original DocRED dataset, and no additional annotation rule that may involve unexamined ethical concerns was introduced. Annotators receive a competitive pay of 100 yuan per hour (more than 4 times the local minimum wage) under the approval of the institute, and both the annotation and discussion stage count towards the paid working time. Annotators are required to read the ACM Code of Ethics before annotating and report any document that violates the code. These documents are removed from the sampled documents. However, there may still be sentences or entities that are from Wikipedia pages with potentially improper content. The possible adoption of such content is not a decision of the authors, and all content in the dataset does not reflect the views or stances of the authors. The resulting re-annotations from the agreement of two expert annotators form a decent approximation of the gold labels, but may still not be the ground truth due to natural error rates. Further use of the dataset should be aware of the limitations and other possible issues, and we are not responsible for issues in further model training processes using our data.

\bibliography{acl}
\bibliographystyle{acl}

\newpage 
\appendix

\section{Source of Recommendations in DocRED}
\label{sec:appendix_source}
According to \citet{DocRED}, Wikidata recommends an average of 19.9 relation instances per document and annotators reserve 57.2\% of these, implying that Wikidata provides an average of 11.4 accurate relations every article. The number of correct relation instances recommended by Wikidata and the RE system does not exceed the number of the instances after the revision phase, which is 12.5 per document. This indicates that at least 90\% of recommendations originate from Wikidata. As a result, when we analyze the bias of DocRED's recommendation, we are mainly discussing the bias of Wikidata.

\section{Entity Bias}
\label{sec:appendix_entity_bias}
\begin{figure}[h]
    \centering
    \includegraphics[width=0.40\textwidth]{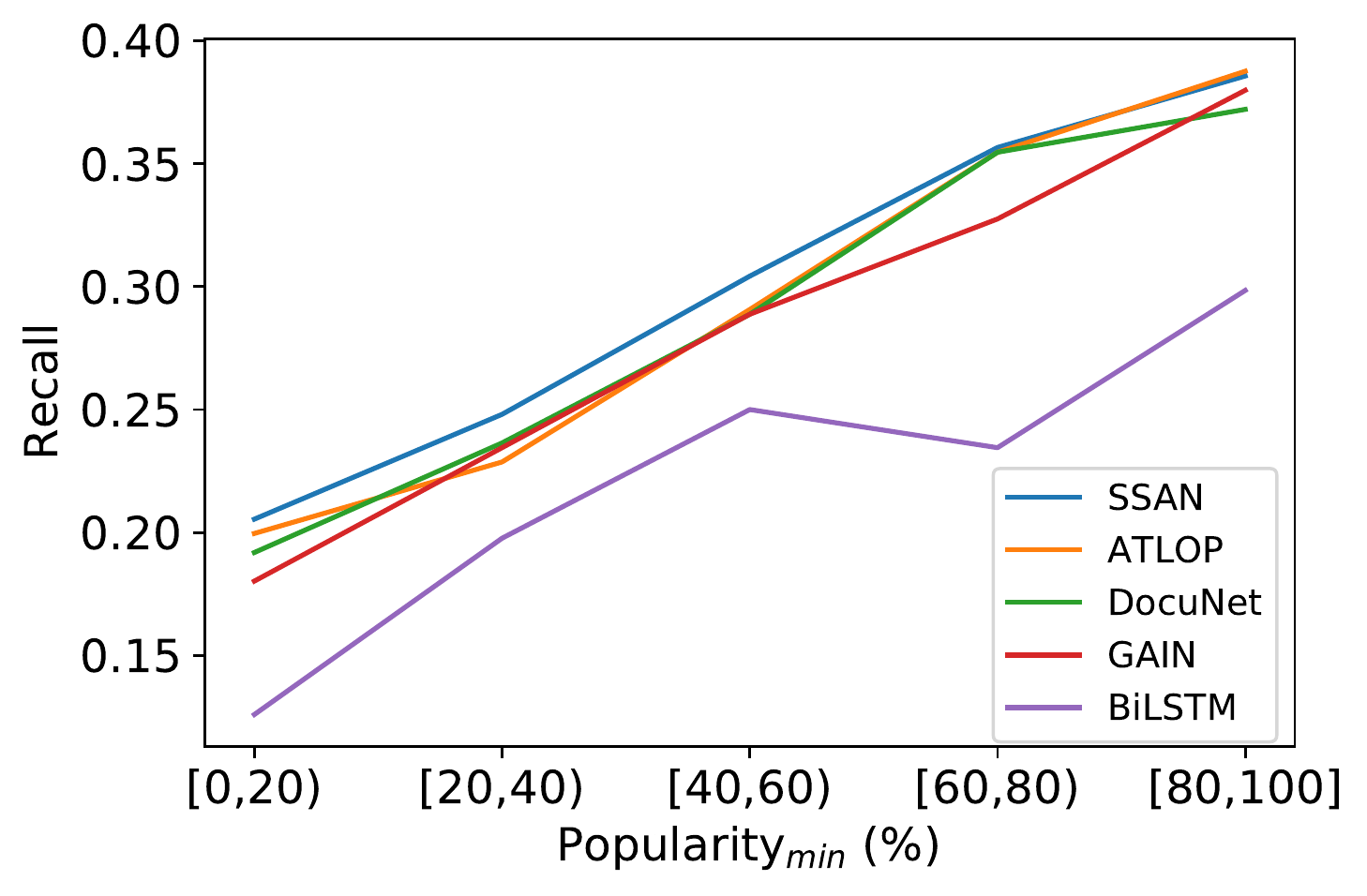}

    \caption{Model's recall in instances with entities of different popularity. We divide all instances in $\mathbf{D_{Scratch}}$ into 5 groups based on the \textit{popularity}$_{min}$ in each instance and we can see that the probability of discovering an instance is positively correlated with its \textit{popularity}$_{min}$.
    }
    \label{fig:entity_bias_min}
    
\end{figure}

\end{document}